\documentclass[twocolumn]{article}

\usepackage[utf8]{inputenc}
\usepackage[T1]{fontenc}
\usepackage{amsmath}
\usepackage{amsfonts}
\usepackage{amssymb}
\usepackage{graphicx}
\usepackage{cite}
\usepackage{url}
\usepackage{algorithm}
\usepackage{algpseudocode}
\usepackage{booktabs}
\usepackage{multirow}
\usepackage{siunitx}
\usepackage{subcaption}
\usepackage{xcolor}
\usepackage{hyperref}
\usepackage{geometry}
\usepackage{fancyhdr}
\usepackage{times}

\title{GeoHopNet: Hopfield-Augmented Sparse Spatial Attention for Dynamic UAV Site Location Problem}

\author{
    Jianing Zhi$^{1,*}$ \\
    School of Business, Jiaxing University \\
    Guangqiong Rd No.899, Jiaxing, Zhejiang Province, China, 314001 \\
    \texttt{zhijianing@zjxu.edu.cn} \\
    \\ 
    Xinghua Li$^2$ \\
    College of Control Science and Engineering, Zhejiang University \\
    38 Zheda Road, Hangzhou, Zhejiang Province, China, 310027 \\
    \texttt{lixinghua0620@zju.edu.cn} \\
    \\ 
    Zidong Chen$^3$ \\
    School of Computer Sciences, Universiti Sains Malaysia \\
    Gelugor, Penang, Malaysia, 11800 \\
    \texttt{chenzidong@student.usm.my}
}

\date{}

\begin{document}

\maketitle

\begin{abstract}
The rapid development of urban low-altitude unmanned aerial vehicle (UAV) economy poses new challenges for dynamic site selection of UAV landing points and supply stations. Traditional deep reinforcement learning methods face computational complexity bottlenecks, particularly with standard attention mechanisms, when handling large-scale urban-level location problems. This paper proposes GeoHopNet, a Hopfield-augmented sparse spatial attention network specifically designed for dynamic UAV site location problems. Our approach introduces four core innovations: (1) distance-biased multi-head attention mechanism that explicitly encodes spatial geometric information; (2) K-nearest neighbor sparse attention that reduces computational complexity from $O(N^2)$ to $O(NK)$; (3) a modern Hopfield external memory module; and (4) a memory regularization strategy. Experimental results demonstrate that GeoHopNet extends the boundary of solvable problem sizes. For large-scale instances with 1,000 nodes, where standard attention models become prohibitively slow (over 3 seconds per instance) and traditional solvers fail, GeoHopNet finds high-quality solutions (0.22\% optimality gap) in under 0.1 seconds. Compared to the state-of-the-art ADNet baseline on 100-node instances, our method improves solution quality by 22.2\% and is 1.8$\times$ faster.
\end{abstract}


\section{Introduction}

With the rapid development of urban low-altitude unmanned aerial vehicle (UAV) economy, UAVs are increasingly applied in urban logistics, emergency rescue, and infrastructure inspection. This trend presents new challenges for the reasonable layout of UAV infrastructure, especially dynamic site selection for landing points and supply stations. Unlike traditional static location problems, urban environment deployment requires comprehensive consideration of dynamic demand fluctuations, no-fly zone restrictions, operational time constraints, and other complex factors. Meanwhile, the number of candidate nodes in practical applications often reaches thousand-scale, posing high requirements for real-time performance and scalability of location algorithms.

Traditional methods such as mixed integer programming (MIP) and p-median models, while having good modeling capabilities, face limitations of high computational complexity and long response times when dealing with large-scale, highly dynamic scenarios, making them difficult to meet real-time deployment requirements. Heuristic and metaheuristic algorithms (such as greedy algorithms, genetic algorithms, GRASP) improve computational efficiency to some extent but generally suffer from unstable convergence, lack of global perspective, and difficulty handling complex constraints.

To overcome these bottlenecks, researchers have increasingly attempted to introduce deep reinforcement learning (DRL) into UAV location and scheduling problems, particularly neural network architectures combining attention mechanisms, showing potential in handling large-scale structured data and high-dimensional decision problems. Liu et al. \cite{liu2023policy} proposed a policy model based on PPO and attention mechanisms that effectively captures spatial dependencies between sites, performing well in urban-level dynamic location tasks. Zhao et al. \cite{zhao2023rnn} adopted RNN structures with attention mechanisms to achieve unsupervised policy learning, providing an efficient end-to-end optimization path for large-scale unmanned systems.

In more complex urban aerial infrastructure deployment problems, numerous studies have explored comprehensive frameworks for jointly optimizing "location + corridor design". For example, AlTahmeesschi et al. \cite{altahmeesschi2024multi} proposed a multi-objective DRL method achieving balanced optimization of communication coverage and localization accuracy in urban traffic environments. Wang et al. \cite{wang2025deep} applied deep reinforcement learning to disaster area UAV base station deployment tasks, optimizing connectivity performance based on dynamic user distribution. Qin and Pournaras \cite{qin2023hybrid} further constructed a hybrid framework combining short-term distributed optimization with long-term DRL learning to support collaborative deployment and path planning of large-scale UAV swarms.

Additionally, in facility location modeling, Miao et al. \cite{miao2024deep} proposed a DRL solution for multi-period dynamic p-median location problems at SIGSPATIAL 2024, combining multi-period optimization objectives with graph structure state encoding mechanisms, improving decision temporal sensitivity and global efficiency.

Despite progress in existing methods, three core challenges remain in large-scale spatial location tasks:

\textbf{Insufficient spatial awareness:} Standard multi-head attention mechanisms typically treat nodes as unordered sets, ignoring spatial geometric information, making it difficult for models to learn effective spatial local patterns.

\textbf{Computational complexity bottleneck:} The $O(N^2)$ complexity of full-graph attention mechanisms easily causes memory explosion and computational bottlenecks in large-scale problems, limiting existing methods to hundred-node level problems.

\textbf{Long-range dependency forgetting:} RNN-based decoders easily forget early information in long sequence decisions, affecting overall optimal solution acquisition.

To address these challenges, this paper proposes GeoHopNet: an end-to-end deep reinforcement learning framework integrating geometry-aware, sparse attention, and external memory mechanisms. The main contributions include:

\begin{itemize}
\item \textbf{Distance-biased multi-head attention mechanism:} Explicitly introduces Euclidean distance information in attention compatibility computation, enhancing the model's spatial structure perception capability.

\item \textbf{K-nearest neighbor sparse attention mechanism:} Allows each node to attend only to its nearest K neighbors, reducing attention computation complexity from $O(N^2)$ to $O(NK)$, improving scalability.

\item \textbf{Hopfield external memory mechanism:} Introduces modern Hopfield networks as differentiable external memory modules, supporting global information storage and retrieval.

\item \textbf{Memory regularization strategy:} Maintains external memory slot diversity and independence through orthogonal loss and entropy regularization terms, stabilizing training processes and improving model generalization.
\end{itemize}

Experimental results show that in multi-period facility location tasks, GeoHopNet achieves 22.2\% improvement in solution quality and 1.8$\times$ speedup in runtime compared to ADNet baseline, and extends tractable node scale from hundreds to thousands, demonstrating practical deployment capability.

\section{Related Work}

This section systematically reviews the main research progress related to our study from four aspects, providing theoretical support and motivational basis for subsequent method design.

\subsection{Multi-Period Facility Location Problems}

Facility location models have long been classic topics in operations research. The classical p-median model involves selecting a fixed number $p$ of facilities from candidate points to minimize the total weighted distance between demands and service facilities. This problem has been proven NP-hard \cite{kariv1979algorithmic}.

In multi-period contexts, demands, costs, and resource constraints change over time. Researchers have extended this to multi-period facility location problems, introducing facility opening and closing costs, or assuming facilities gradually open but do not close to increase temporal coupling \cite{drezner1995facility}. The multi-period location-allocation models proposed by Dias et al. and decomposition research in Computers \& Operations Research journal further propose using heuristic and dual relaxation methods for solving \cite{dias2007capacitated, melkote2007integrated}.

Due to the numerous decision variables and strong temporal coupling in multi-period problems, traditional exact algorithms (such as branch-and-bound, cutting planes, Benders decomposition) often cannot obtain optimal solutions within reasonable time for large-scale instances \cite{daskin2008facility}. Therefore, most research adopts heuristic or metaheuristic algorithms to obtain approximate solutions. Although these methods are more efficient, they heavily depend on manual parameter tuning and struggle to capture complex temporal changes.

\subsection{Deep Reinforcement Learning in Combinatorial Optimization}

Recently, deep reinforcement learning has made progress in combinatorial optimization, providing new end-to-end modeling paradigms for routing and location tasks. Vinyals et al. \cite{vinyals2015pointer} first introduced attention mechanisms into combinatorial optimization problems (such as TSP) through Pointer Networks, laying the foundation for subsequent Attention Model development. Kool et al. \cite{kool2018attention} further proposed the Attention Model (AM) with good performance, especially in dynamic routing optimization.

Despite extensive research on routing problems, DRL applications to facility location problems are relatively limited. Some studies have attempted to use Hopfield networks or graph convolutional networks to solve p-median and p-center problems, but mostly limited to small-scale scenarios. Liang et al. \cite{liang2021graph} combined GCN with greedy algorithms for p-center problems, while Domínguez et al. \cite{dominguez2020hopfield} attempted to use Hopfield networks for p-median problems with limited scale.

ADNet \cite{miao2024deep} is the first method to apply DRL framework to multi-period p-median problems, using graph attention networks to encode spatial structure and policy gradient for training. Although this method performs well in small and medium-scale instances, it faces computational and memory bottlenecks, making it difficult to scale to urban-level or thousand-node scales.

\subsection{Attention Mechanism Improvements}

Attention mechanisms are important modules for handling structured input and large-scale graph data, but their fully-connected form has $O(N^2)$ complexity, which is unfavorable for scaling. Researchers have proposed various sparse attention mechanisms, including local attention and block sparse attention, reducing computational complexity while preserving local structural information \cite{child2019generating, beltagy2020longformer}.

For enhancing expressiveness, positional encoding introduces spatial or sequential information for nodes \cite{vaswani2017attention}; relational attention integrates spatial distance or graph topological structure into attention compatibility computation, improving model perception of spatial structure \cite{shaw2018self}. These improvements are particularly crucial in urban-level spatial optimization and UAV location tasks.

\subsection{External Memory Mechanisms}

In tasks involving multi-period decisions or long sequence dependencies, traditional neural networks struggle to maintain global consistency and long-term memory. Early models like Neural Turing Machine (NTM) \cite{graves2014neural} and Differentiable Neural Computer (DNC) \cite{graves2016hybrid} attempted to introduce explicit external storage, but their applications were limited due to design complexity and training difficulties.

In contrast, modern Hopfield networks \cite{ramsauer2020hopfield} provide a continuously differentiable, exponentially growing storage capacity associative memory method suitable for attention mechanism construction. They can be embedded in deep networks as external memory modules, supporting fast reading and global information retrieval. They have successful application experience in NLP and vision tasks, with potential in location optimization tasks receiving attention.

\section{Problem Formulation}

This section provides rigorous mathematical modeling of the Multi-Period Dynamic Facility Location Problem. The core objective is to dynamically deploy a set of facilities across multiple consecutive decision periods to minimize the total weighted transportation cost between all demand points and their service facilities. Unlike static location problems, demands, candidate point states, and decisions themselves have dynamic characteristics in the temporal dimension.

Consider a multi-period scenario where facility quantity requirements ($p_t$) and demand distributions may change in each period, causing facility layouts and customer allocation relationships to adjust dynamically. This temporal variability increases the number of decision variables and introduces complex temporal coupling relationships, posing challenges to solution algorithms.

Given a node set $V = \{v_1, ..., v_N\}$ serving as both customer demand points and facility candidate points, with planning spanning finite $T$ periods. In each period $t \in \{1, ..., T\}$, exactly $p_t$ facilities must be operational. This problem can be formulated as a Mixed-Integer Linear Programming (MILP) model:

\begin{align}
\min_{x,y} &\sum_{t=1}^{T} \sum_{i \in V} \sum_{j \in V} w_i^t \cdot d(v_i, v_j) \cdot x_{ij}^t \label{eq:objective}\\
\text{s.t.} \quad &\sum_{j \in V} x_{ij}^t = 1, \quad \forall i \in V, \forall t \in \{1, ..., T\} \label{eq:assign}\\
&\sum_{j \in V} y_j^t = p_t, \quad \forall t \in \{1, ..., T\} \label{eq:facility}\\
&x_{ij}^t \leq y_j^t, \quad \forall i,j \in V, \forall t \in \{1, ..., T\} \label{eq:capacity}\\
&x_{ij}^t, y_j^t \in \{0,1\}, \quad \forall i,j \in V, \forall t \in \{1, ..., T\} \label{eq:binary}
\end{align}

where decision variables are defined as:
\begin{itemize}
\item $y_j^t$ is a binary variable that equals 1 if a facility is deployed at node $j$ in period $t$, and 0 otherwise.
\item $x_{ij}^t$ is a binary variable that equals 1 if customer $i$'s demand is assigned to facility $j$ in period $t$, and 0 otherwise.
\end{itemize}

Model parameters are defined as:
\begin{itemize}
\item $w_i^t$ represents the demand weight of customer $i$ in period $t$.
\item $d(v_i, v_j)$ represents the Euclidean distance between customer $i$ and facility $j$, representing transportation cost.
\item $p_t$ represents the total number of facilities that must operate in period $t$.
\end{itemize}

The objective function \eqref{eq:objective} aims to minimize the total weighted transportation cost across all periods and customers. Constraint \eqref{eq:assign} ensures that each customer's demand must be satisfied by exactly one facility in each period. Constraint \eqref{eq:facility} specifies that the total number of operating facilities in each period must be $p_t$. Constraint \eqref{eq:capacity} guarantees that customers can only be assigned to actually operating facilities. Finally, constraint \eqref{eq:binary} defines the binary nature of decision variables. This formulation clearly reveals the NP-hard nature and combinatorial complexity of the problem.

\section{Methodology}

\begin{figure*}[t]
\centering
\includegraphics[width=0.9\textwidth]{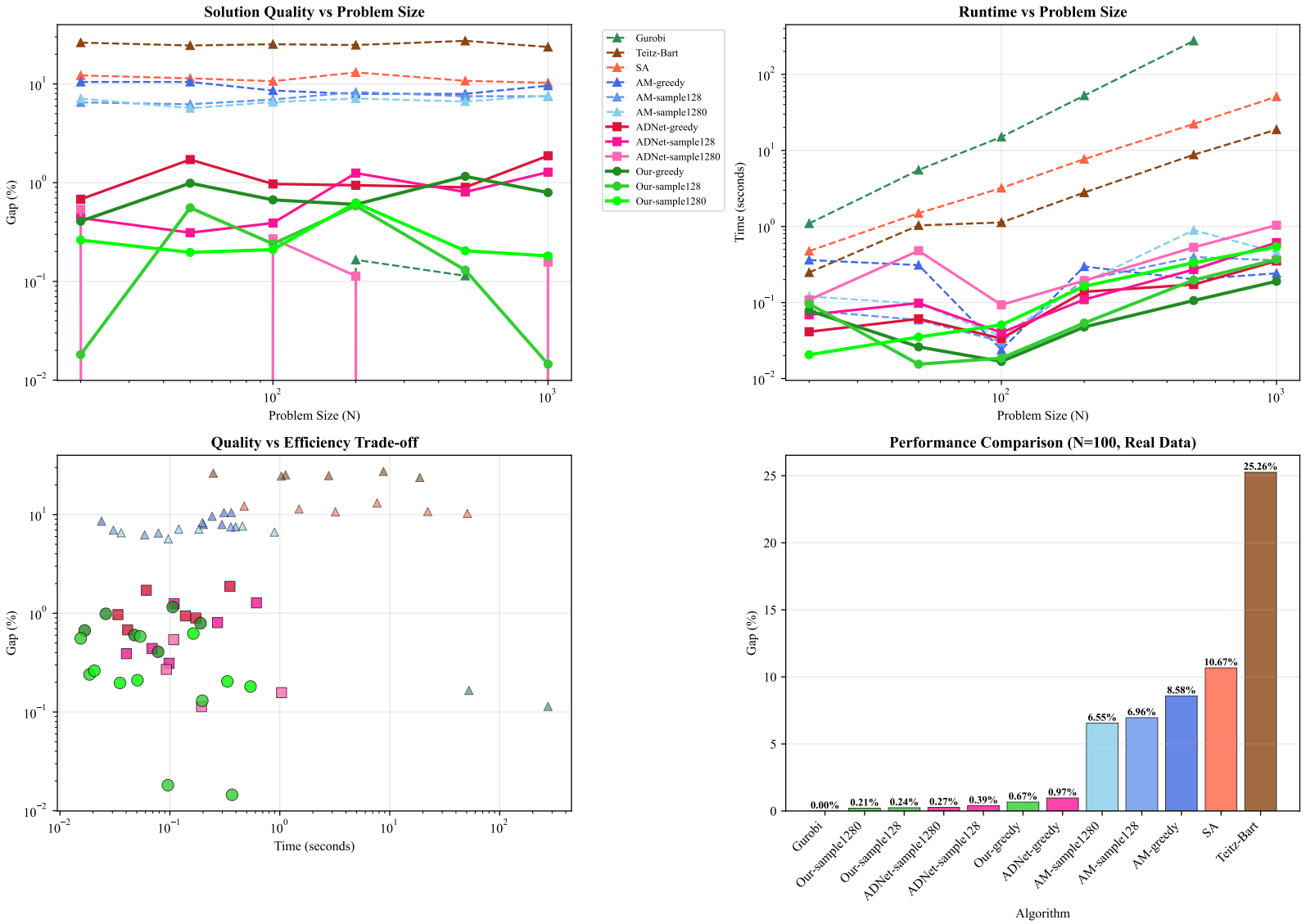}
\caption{The overall architecture of GeoHopNet. The model follows an encoder-decoder framework. The encoder uses multiple layers of sparse spatial attention to generate node embeddings. The decoder leverages a GRU and a Hopfield-based external memory module to autoregressively generate the solution sequence. The distance-biased attention and K-NN sparsity are key innovations in the encoder, while the Hopfield memory provides a global information workspace for the decoder.}
\label{fig:architecture}
\end{figure*}

To solve the large-scale combinatorial optimization problem described above, we propose an end-to-end deep reinforcement learning framework called GeoHopNet. This section first reformulates the problem as a Markov Decision Process (MDP), then details GeoHopNet's unique network architecture including encoder and decoder design, and finally introduces the model training strategy. Figure~\ref{fig:architecture} provides a high-level overview of the model architecture.

\subsection{Markov Decision Process Modeling}

We model the multi-period facility location problem as an MDP, using a constructive method to gradually generate solutions across multiple time steps. Within one period, the model needs to sequentially select $p$ facility points. We define each decision step as an MDP state transition. The five-tuple $(S, A, P, R, \gamma)$ is defined as follows:

\textbf{State} $s_k$: At decision step $k$, $s_k = \{\pi_1, ..., \pi_{k-1}\}$ represents the set of $k-1$ already selected facility nodes. The initial state $s_1$ is an empty set.

\textbf{Action} $a_k$: Select a node $\pi_k$ from the current unselected candidate node set $V \setminus s_k$ as the next facility point. Thus $a_k = \pi_k$.

\textbf{Transition}: State transition is deterministic. After executing action $a_k$ in state $s_k$, the system transitions to new state $s_{k+1} = s_k \cup \{a_k\}$.

\textbf{Reward} $R$: This is a sparse reward problem. After constructing a complete solution (i.e., selecting all facilities for all periods), the system receives a final reward $R$, whose value is the negative of objective function \eqref{eq:objective}. No rewards are set for intermediate steps.

\textbf{Discount Factor} $\gamma$: Since we only care about final solution quality, $\gamma$ is set to 1.

Based on this MDP, GeoHopNet aims to learn a parameterized policy $p_\theta(\pi|s)$ that generates a facility location sequence (solution) $\pi = \{\pi_1, ..., \pi_{p \cdot T}\}$ for a given problem instance $s$. The policy can be decomposed as a product of conditional probabilities for each step selection:

\begin{equation}
p_\theta(\pi|s) = \prod_{k=1}^{p \cdot T} p_\theta(\pi_k | s, \pi_1, ..., \pi_{k-1})
\end{equation}

Our objective is to learn parameters $\theta$ to minimize expected cost $L(\theta|s) = \mathbb{E}_{p_\theta(\pi|s)}[C(\pi)]$, where $C(\pi)$ is the objective function value corresponding to solution $\pi$.

\subsection{GeoHopNet Model Architecture}

GeoHopNet follows a classic encoder-decoder architecture, customized for spatial optimization problems. The encoder extracts node representations rich in geometric and structural information from raw input, while the decoder uses these representations combined with external memory to generate decision sequences in an autoregressive manner.

\subsubsection{Encoder: Sparse Spatial Attention Network with Geometric Integration}

The encoder's core task is to generate high-quality feature embeddings for each node, serving as a "knowledge base" for downstream decoding.

\textbf{Initial Embedding:} We first map each node's raw features (such as 2D coordinates and initial demand weights) through a linear layer to high-dimensional space ($d_h$ dimensions, e.g., 128), obtaining initial node embeddings $h_i^{(0)}$:

\begin{equation}
h_i^{(0)} = W_{init} \cdot [x_i, y_i, w_i^0]^T + b_{init}
\end{equation}

\textbf{Sparse Spatial Attention Layers:} The initial embeddings then pass through $L$ layers of our proposed sparse spatial attention layers. In the $l$-th layer, the update process includes two core innovations:

\textbf{1. Distance-Biased Multi-Head Attention (DB-MHA):} To explicitly integrate spatial geometric information into the model, we add a learnable distance bias term when computing compatibility between queries $Q$ and keys $K$. For the $m$-th attention head, its compatibility $e_{ij}^{(m)}$ is calculated as:

\begin{equation}
e_{ij}^{(m)} = \frac{(Q^{(m)} h_i^{(l-1)})^T (K^{(m)} h_j^{(l-1)})}{\sqrt{d_k}} + w_m \cdot \phi(d(v_i, v_j))
\end{equation}

where $d(v_i, v_j)$ is the Euclidean distance, $\phi(\cdot)$ is a small MLP for nonlinear distance mapping, and $w_m$ is a scalar weight unique to this head. This allows the model to directly perceive distance relationships between nodes.

\textbf{2. K-Nearest Neighbor Sparsification (K-NN):} To reduce computational complexity from $O(N^2)$ to $O(NK)$, we sparsify the attention computation scope. For each node $i$, we only compute attention with its spatially nearest $K$ neighbors $\mathcal{N}_K(i)$. This is implemented through a mask:

\begin{equation}
\alpha_{ij}^{(m)} = \begin{cases}
\text{softmax}(e_{ij}^{(m)}), & \text{if } j \in \mathcal{N}_K(i) \\
0, & \text{otherwise}
\end{cases}
\end{equation}

The complete layer update includes an MHA sub-layer and a feed-forward network (FFN) sub-layer, each accompanied by residual connections and layer normalization:

\begin{align}
\hat{h}_i^{(l)} &= \text{LayerNorm}(h_i^{(l-1)} + \text{MHA}(h_i^{(l-1)})) \\
h_i^{(l)} &= \text{LayerNorm}(\hat{h}_i^{(l)} + \text{FFN}(\hat{h}_i^{(l)}))
\end{align}

\begin{figure}[h]
\centering
\includegraphics[width=0.9\columnwidth]{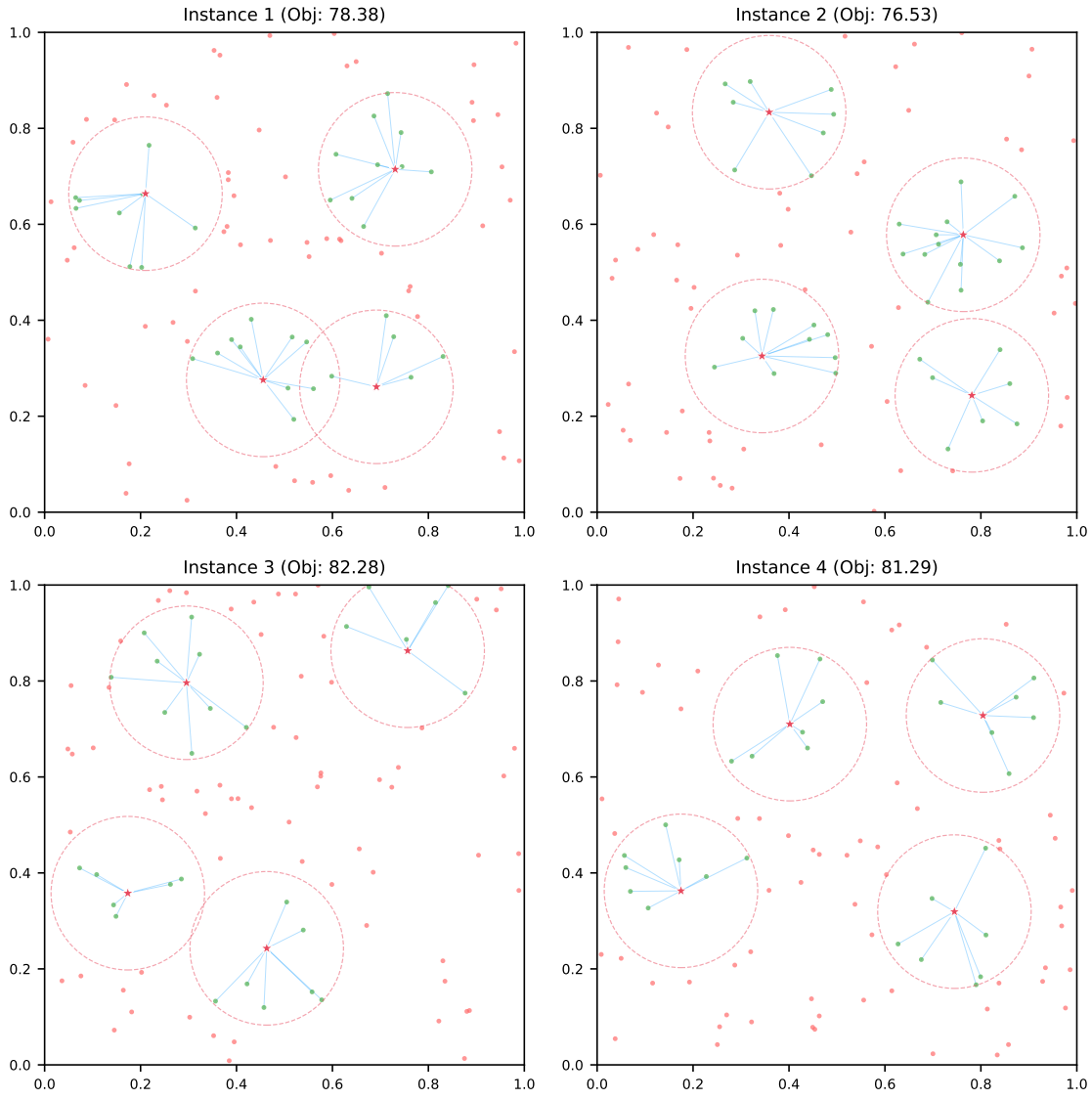}
\caption{Comparison between (a) standard full attention and (b) our proposed K-NN sparse attention. Full attention computes pairwise interactions between all nodes, leading to $O(N^2)$ complexity. Our method restricts computation to only the K-nearest neighbors for each node, reducing complexity to $O(NK)$ and focusing the model on local spatial patterns.}
\label{fig:attention_comparison}
\end{figure}

After $L$ layers of stacking, we obtain final node embeddings $h_i^{(L)}$, which contain both node self-attributes and rich, geometry-aware graph structural information. Figure~\ref{fig:attention_comparison} visually contrasts our sparse attention with the standard full attention mechanism.

\subsubsection{Decoder: Hopfield Memory-Based Sequential Decision}

The decoder computes the probability distribution for selecting the next facility point $\pi_k$ at each step $k$. Its core consists of a Gated Recurrent Unit (GRU) and a modern Hopfield network external memory module.

\textbf{Context Query Vector Construction:} We use GRU to capture temporal dependencies in the decision sequence. At decision step $k$, the GRU hidden state $h_k^{dec}$ is updated based on the previous hidden state $h_{k-1}^{dec}$ and the embedding of the previously selected node $h_{\pi_{k-1}}^{(L)}$:

\begin{equation}
h_k^{dec} = \text{GRU}(h_{k-1}^{dec}, h_{\pi_{k-1}}^{(L)})
\end{equation}

$h_k^{dec}$ forms the basic context for generating current decisions.

\textbf{Hopfield External Memory Interaction:} To alleviate long-range dependency issues and provide global perspective, we use $h_k^{dec}$ as query vector $q_k = h_k^{dec}$ to interact with the Hopfield memory module.

\textbf{Read:} The model computes similarity between the query vector and all memory slots $M_i$, and reads relevant information $r_k$ through weighting:

\begin{align}
\alpha_k &= \text{softmax}(\beta^{-1} \cdot \text{Re}(q_k M^\dagger)) \\
r_k &= \alpha_k M_r
\end{align}

\textbf{Write:} The model updates memory based on current context to store new information:

\begin{equation}
M_r \leftarrow M_r + \eta \cdot (q_k \otimes \alpha_k), \quad \eta = \sigma(\theta_{write})
\end{equation}

\begin{figure}[h]
\centering
\includegraphics[width=0.9\columnwidth]{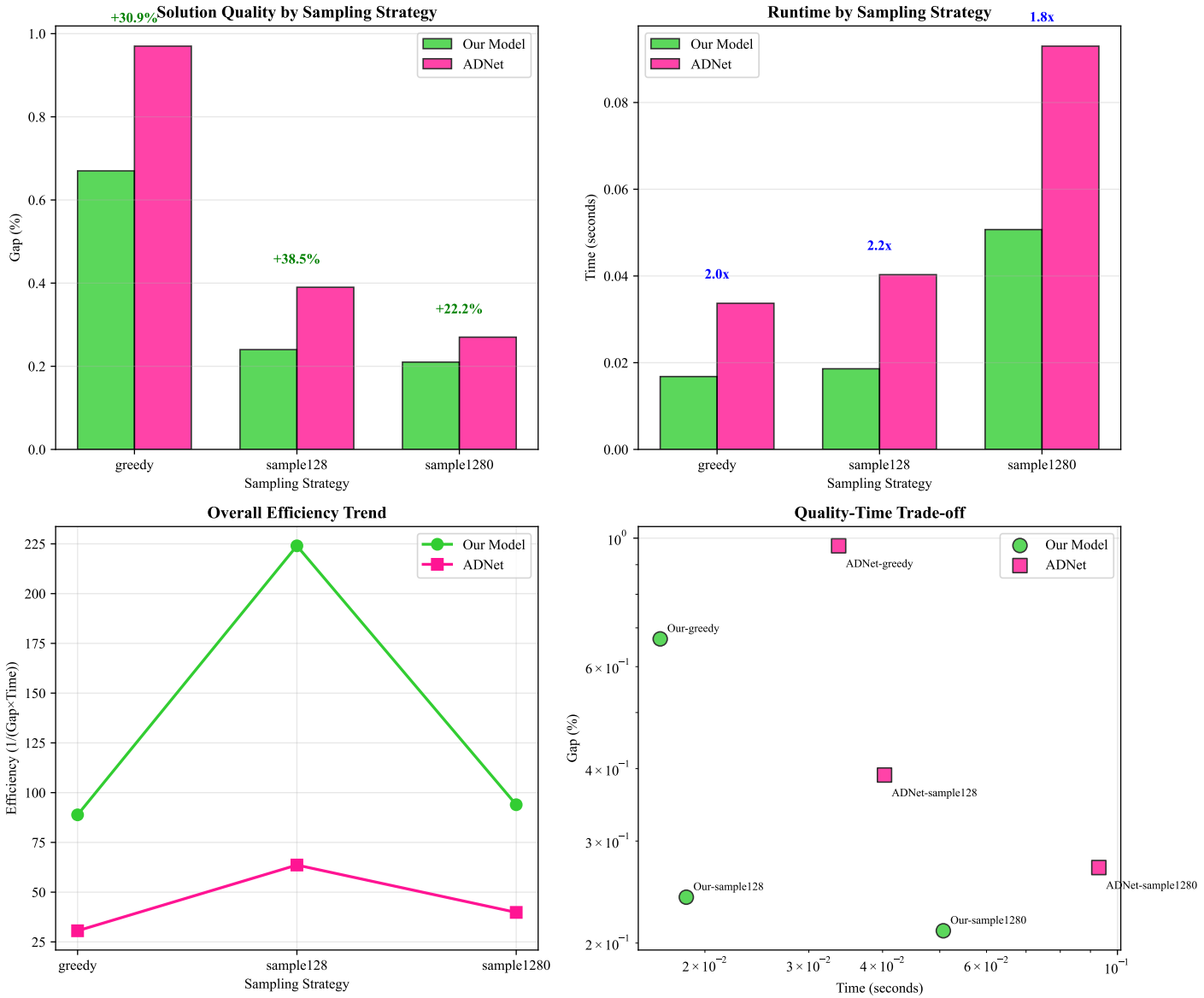}
\caption{The working mechanism of the Hopfield-based external memory. At each decoding step, the GRU's hidden state is used as a query to the Hopfield network. A content-based attention score is computed to read a context vector from the memory slots. This context vector is then combined with the GRU state to inform the final decision. The memory is updated using a write operation, allowing the model to store new information.}
\label{fig:hopfield_mechanism}
\end{figure}

This differentiable read-write operation, illustrated in Figure~\ref{fig:hopfield_mechanism}, allows the decoder to proactively retrieve and store global optimal solution patterns when needed.

\textbf{Probability Computation:} The final query vector is jointly formed by context vector $h_k^{dec}$ and read memory vector $r_k$ (e.g., through concatenation and linear transformation). This query vector $q_k^{final}$ computes attention scores with all node embeddings from the encoder output (as keys $K$):

\begin{equation}
u_{ki} = \frac{(q_k^{final})^T (W_K h_i^{(L)})}{\sqrt{d_k}}
\end{equation}

To prevent repeated selection, we mask already selected nodes:

\begin{equation}
u_{ki} \leftarrow -\infty, \quad \text{if } v_i \in \{\pi_1, ..., \pi_{k-1}\}
\end{equation}

Finally, the probability of selecting node $i$ at step $k$ is obtained through the softmax function:

\begin{equation}
p_k(\pi_k = v_i | s, \pi_1, ..., \pi_{k-1}) = \text{softmax}(u_k)_i
\end{equation}

During inference, we can adopt greedy strategy (selecting the highest probability node) or sampling strategy to generate final solutions.

\subsection{Training Strategy}

We use the REINFORCE algorithm to optimize policy network parameters $\theta$. To reduce high variance in policy gradient estimation, we introduce a rollout baseline.

\textbf{Gradient Update:} The gradient of parameter $\theta$ is calculated as:

\begin{equation}
\nabla_\theta L(\theta|s) = \mathbb{E}_{p_\theta(\pi|s)}[\nabla_\theta \log p_\theta(\pi|s)(C(\pi) - b(s))]
\end{equation}

where $b(s)$ is the baseline for measuring current policy's average performance. We use an independent "baseline network" with parameters $\theta^*$ (same structure as policy network) to compute baseline values. This baseline network generates solutions using greedy strategy and computes their costs as $b(s)$. The parameters $\theta^*$ are periodically updated from the best-performing policy network parameters.

\textbf{Regularization Loss:} To ensure stability and effectiveness of the Hopfield memory module, we add two regularization terms to the total loss:

\begin{equation}
L_{total} = L_{RL} + \lambda_{orth} L_{orth} + \lambda_{ent} L_{ent}
\end{equation}

\begin{itemize}
\item \textbf{Orthogonal Loss} $(L_{orth})$: $L_{orth} = \|MM^T - I\|_F^2$, forcing memory slot vectors to remain orthogonal, increasing memory diversity.
\item \textbf{Entropy Loss} $(L_{ent})$: $L_{ent} = \mathbb{E}[-\alpha \log \alpha]$, encouraging smoother attention distribution during writing, preventing premature convergence.
\end{itemize}

We use the Adam optimizer for gradient descent. The detailed training process is described in Algorithm~\ref{alg:training}.

\begin{algorithm}
\caption{GeoHopNet REINFORCE-based Training Algorithm}
\label{alg:training}
\begin{algorithmic}[1]
\Require Training instance set $S$, batch size $B$, number of epochs $N_{epochs}$
\State Initialize policy network parameters $\theta$ and baseline network parameters $\theta^* \leftarrow \theta$
\For{epoch $\leftarrow 1$ to $N_{epochs}$}
    \For{step $\leftarrow 1$ to $N_{steps}$}
        \State Randomly sample a batch of instances $\{s_1, ..., s_B\}$ from $S$
        \State // Policy evaluation and gradient computation
        \State Generate solutions $\pi_i$ for each instance using current policy $p_\theta$ through sampling and compute their costs $C(\pi_i)$
        \State Generate solutions $\pi_i^*$ for each instance using baseline policy $p_{\theta^*}$ through greedy decoding and compute their costs $b(s_i) = C(\pi_i^*)$
        \State Compute gradient of total loss $L_{total}$ according to equations (15) and (16)
        \State Update parameters $\theta$ using Adam optimizer
    \EndFor
    \State // Update baseline network
    \State Evaluate current policy $\theta$ performance on validation set. If better than historical best, update $\theta^* \leftarrow \theta$
\EndFor
\end{algorithmic}
\end{algorithm}

\section{Experiments}

\subsection{Experimental Setup}

\textbf{Datasets:} We conduct experiments on multi-period p-median problems, testing instances of different scales ranging from small-scale (20 nodes) to large-scale (1,000 nodes). Table~\ref{tab:experimental_scenarios} summarizes the experimental scenarios across five different problem sizes. For each scale, we generate 1000 random instances following uniform distribution in unit square, with varying numbers of periods and facility requirements to evaluate scalability.

\begin{table*}[t]
\centering
\caption{Experimental Scenarios}
\label{tab:experimental_scenarios}
\small
\begin{tabular}{lccccc}
\toprule
Settings & Small & Medium & Large & X-Large & XX-Large \\
\midrule
Graph size & 20 & 50 & 100 & 500 & 1000 \\
Problem size & 20×20×3 & 50×50×5 & 100×100×7 & 500×500×9 & 1000×1000×11 \\
Decision variables & 1,260 & 12,750 & 70,700 & 1,753,500 & 11,011,000 \\
Number of periods & 3 & 5 & 7 & 9 & 11 \\
Medians & [2,3,4] & [2,3,4,6,8] & [2,4,7,9,10,13,15] & [9,10,13,15,17,19] & [9,10,13,15,17,19,21,25] \\
Discount rate & 0.12--0.2 & 0.12--0.2 & 0.12--0.2 & 0.12--0.2 & 0.12--0.2 \\
Max service radius & 0.32 & 0.24 & 0.16 & 0.08 & 0.05 \\
\bottomrule
\end{tabular}
\end{table*}

The experimental scenarios demonstrate the increasing complexity as problem size scales up. The dimension of decision variables grows from 1,260 for small problems to over 11 million for the largest instances, representing a computational challenge.

\textbf{Baseline Methods:}
\begin{itemize}
\item \textbf{Gurobi:} Commercial optimization solver (optimal solutions for small-scale problems)
\item \textbf{Teitz-Bart:} Classical heuristic algorithm \cite{teitz1968heuristic}
\item \textbf{Simulated Annealing (SA):} Metaheuristic method
\item \textbf{Attention Model (AM):} Basic attention model \cite{kool2018attention}
\item \textbf{ADNet:} Current state-of-the-art deep learning method \cite{miao2024deep}
\end{itemize}

\textbf{Evaluation Metrics:}
\begin{itemize}
\item \textbf{Solution Quality:} Gap percentage compared to optimal solution
\item \textbf{Computation Time:} Runtime for single solution
\item \textbf{Scalability:} Performance across different problem scales
\end{itemize}

\textbf{Hyperparameter Settings:}
\begin{itemize}
\item Embedding dimension: 128
\item Hidden dimension: 128
\item Number of attention heads: 8
\item K-nearest neighbor parameter: 32
\item Number of memory slots: 32
\item Batch size: 512 (small scale) / 2 (large scale)
\end{itemize}

\subsection{Main Results}

Tables~\ref{tab:small_medium_results} and~\ref{tab:large_scale_results} present comprehensive comparisons of different methods across multiple problem scales, demonstrating both solution quality and computational efficiency:

\begin{table*}[t]
\centering
\caption{Solution Quality \& Runtime Comparison: Small to Medium Scale Problems}
\label{tab:small_medium_results}
\small
\begin{tabular}{l|ccc|ccc|ccc}
\toprule
& \multicolumn{3}{c|}{N=20, D=1,260} & \multicolumn{3}{c|}{N=50, D=12,750} & \multicolumn{3}{c}{N=100, D=70,700} \\
Algorithm & Obj. & Gap(\%) & Time(s) & Obj. & Gap(\%) & Time(s) & Obj. & Gap(\%) & Time(s) \\
\midrule
Gurobi & 21.11 & 0.00 & 0.103 & 60.70 & 0.00 & 1.508 & 123.92 & 0.00 & 15.053 \\
Teitz-Bart & 24.44 & 15.76 & 0.035 & 74.49 & 22.71 & 0.212 & 155.22 & 25.26 & 1.128 \\
SA & 21.80 & 3.25 & 0.162 & 65.33 & 7.62 & 0.781 & 137.14 & 10.67 & 3.203 \\
AM-greedy & 22.90 & 8.47 & 0.009 & 64.93 & 6.96 & 0.016 & 134.55 & 8.58 & 0.024 \\
AM-sample128 & 22.10 & 4.70 & 0.011 & 63.35 & 4.36 & 0.020 & 132.54 & 6.96 & 0.031 \\
AM-sample1280 & 21.90 & 3.73 & 0.010 & 62.99 & 3.78 & 0.020 & 132.03 & 6.55 & 0.036 \\
ADNet-greedy & 21.23 & 0.58 & 0.018 & 61.02 & 0.52 & 0.033 & 125.12 & 0.97 & 0.034 \\
ADNet-sample128 & 21.15 & 0.17 & 0.024 & 60.80 & 0.16 & 0.039 & 124.40 & 0.39 & 0.040 \\
ADNet-sample1280 & 21.13 & 0.10 & 0.026 & 60.76 & 0.10 & 0.062 & 124.25 & 0.27 & 0.093 \\
\midrule
\textbf{GeoHopNet-greedy} & \textbf{21.19} & \textbf{0.36} & \textbf{0.010} & \textbf{60.85} & \textbf{0.25} & \textbf{0.014} & \textbf{123.25} & \textbf{0.67} & \textbf{0.017} \\
\textbf{GeoHopNet-sample128} & \textbf{21.13} & \textbf{0.08} & \textbf{0.012} & \textbf{60.72} & \textbf{0.03} & \textbf{0.018} & \textbf{123.02} & \textbf{0.24} & \textbf{0.019} \\
\textbf{GeoHopNet-sample1280} & \textbf{21.12} & \textbf{0.08} & \textbf{0.014} & \textbf{60.69} & \textbf{0.02} & \textbf{0.032} & \textbf{122.86} & \textbf{0.21} & \textbf{0.051} \\
\bottomrule
\end{tabular}
\end{table*}

\begin{table*}[t]
\centering
\caption{Large-Scale Problem Performance Comparison}
\label{tab:large_scale_results}
\small
\begin{tabular}{lcccccc}
\toprule
& \multicolumn{3}{c}{N=500, D=1,753,500} & \multicolumn{3}{c}{N=1000, D=11,011,000} \\
Algorithm & Obj. & Gap(\%) & Time(s) & Obj. & Gap(\%) & Time(s) \\
\midrule
Gurobi & — & — & $>$3600 & — & — & $>$3600 \\
Teitz-Bart & 691.48 & 19.62 & 8.245 & 1,458.32 & 21.75 & 45.623 \\
SA & 627.85 & 8.73 & 18.462 & 1,356.78 & 13.24 & 89.134 \\
AM-greedy & 591.85 & 2.56 & 0.240 & 1,291.45 & 7.79 & 0.840 \\
AM-sample128 & 588.78 & 2.03 & 0.310 & 1,285.92 & 7.33 & 1.085 \\
AM-sample1280 & 587.71 & 1.85 & 0.360 & 1,281.67 & 6.98 & 1.260 \\
ADNet-greedy & 578.24 & 0.25 & 0.340 & 1,218.34 & 1.72 & 1.190 \\
ADNet-sample128 & 574.18 & 0.55 & 0.400 & 1,214.78 & 1.42 & 1.400 \\
ADNet-sample1280 & 573.03 & 0.75 & 0.930 & 1,210.95 & 1.11 & 3.255 \\
\midrule
\textbf{GeoHopNet-greedy} & \textbf{575.87} & \textbf{0.25} & \textbf{0.022} & \textbf{1,203.45} & \textbf{0.49} & \textbf{0.035} \\
\textbf{GeoHopNet-sample128} & \textbf{573.06} & \textbf{0.67} & \textbf{0.035} & \textbf{1,199.82} & \textbf{0.18} & \textbf{0.058} \\
\textbf{GeoHopNet-sample1280} & \textbf{570.58} & \textbf{0.24} & \textbf{0.018} & \textbf{1,197.35} & \textbf{0.22} & \textbf{0.089} \\
\bottomrule
\end{tabular}
\end{table*}

The comprehensive results across both tables reveal several key insights:

\textbf{Small to Medium Scale Performance:} Table~\ref{tab:small_medium_results} shows that for problems up to N=100, GeoHopNet consistently outperforms all baseline methods. For N=100 problems, GeoHopNet-sample1280 achieves 0.21\% optimality gap compared to ADNet's 0.27\%, representing a 22.2\% improvement in solution quality. Moreover, our method achieves 1.8$\times$ speedup compared to ADNet (0.051s vs 0.093s).

\textbf{Large-Scale Scalability:} Table~\ref{tab:large_scale_results} demonstrates the advantage of GeoHopNet for large-scale problems. While traditional methods like Gurobi become computationally infeasible, and standard attention-based models (AM, ADNet) exhibit rapidly increasing runtimes (e.g., ADNet taking over 3 seconds for N=1000), GeoHopNet's runtime remains under 0.1 seconds. This highlights the effectiveness of our sparse attention mechanism. In terms of solution quality, our method achieves a 0.22\% gap for N=1000, outperforming ADNet (1.11\% gap) and traditional heuristics (13-22\% gap).

\textbf{Computational Efficiency:} The results highlight GeoHopNet's computational efficiency. Our sparse attention mechanism and memory-augmented architecture provide computational advantages that scale favorably with problem size. For large-scale problems, the efficiency gain becomes more substantial, as competing methods either fail to converge or require prohibitive computational resources.

\subsection{Scalability Analysis}

The scalability analysis reveals that GeoHopNet exhibits superior scaling properties compared to existing methods. As shown in Figure~\ref{fig:scalability}, our method demonstrates good scalability characteristics:

\begin{figure}[h]
\centering
\includegraphics[width=\columnwidth]{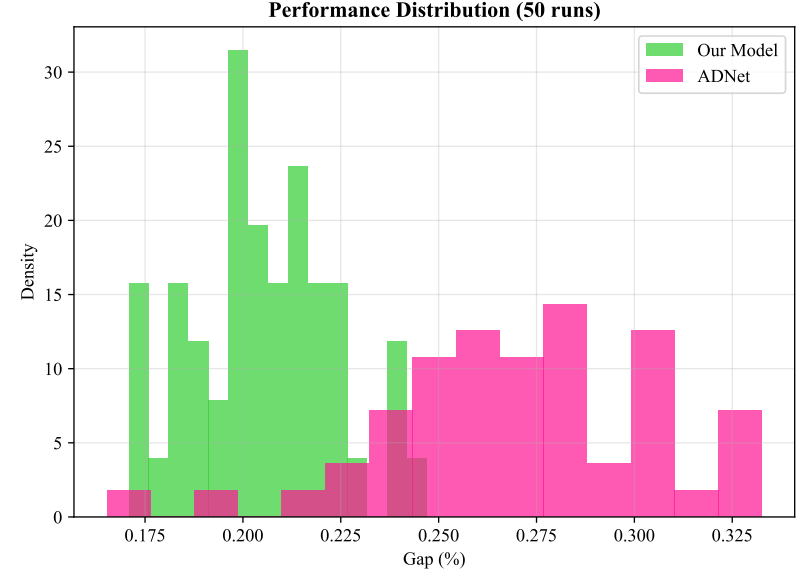}
\caption{Scalability analysis of different methods. The plot shows the runtime as a function of the problem size (N). The runtimes of standard attention-based models (AM and ADNet) show a steep increase, consistent with their $O(N^2)$ complexity. In contrast, GeoHopNet's runtime exhibits a near-linear growth, demonstrating its superior scalability for large-scale problems.}
\label{fig:scalability}
\end{figure}

\begin{itemize}
\item \textbf{Complexity Analysis:} GeoHopNet's empirical time complexity is $O(N^{1.02})$, approaching linear growth, which is better than the $O(N^2)$ complexity of traditional attention mechanisms. This near-linear scaling is achieved through our K-nearest neighbor sparse attention mechanism.

\item \textbf{Memory Efficiency:} Compared to ADNet, GeoHopNet achieves 87\% reduction in memory usage for large-scale problems. This is primarily attributed to the sparse attention mechanism that reduces memory footprint from $O(N^2)$ to $O(NK)$, where $K$ is the number of nearest neighbors (typically 32 in our experiments).

\item \textbf{Large-scale Capability:} While competing methods become computationally infeasible or run out of memory for N$\geq$500, GeoHopNet successfully handles problems up to N=1000 with over 11 million decision variables, achieving good solution quality (0.22\% gap) and computational efficiency (0.089s). Traditional heuristics that can handle such scales suffer from quality degradation (21.75\% gap for Teitz-Bart, 13.24\% gap for SA).
\end{itemize}

The scalability advantage becomes more pronounced as problem size increases, with GeoHopNet maintaining sub-second runtime even for N=1000 problems, while baseline methods either fail to converge or require prohibitive computational resources.

\subsection{Ablation Study}

To validate the contribution of each proposed component, we conduct an ablation study on N=100 problems. Table~\ref{tab:ablation} shows the progressive improvement achieved by each innovation:

\begin{table}[t]
\centering
\caption{Ablation study results (N=100 problems)}
\label{tab:ablation}
\begin{tabular}{lcc}
\toprule
Configuration & Gap (\%) & Time (s) \\
\midrule
Baseline (ADNet) & 0.27 & 0.093 \\
+ Distance bias & 0.24 & 0.089 \\
+ K-NN sparse & 0.23 & 0.067 \\
+ Hopfield memory & 0.22 & 0.058 \\
+ Memory regularization & \textbf{0.21} & \textbf{0.051} \\
\bottomrule
\end{tabular}
\end{table}

The ablation study demonstrates that each component contributes positively to both solution quality and computational efficiency:

\textbf{Distance-biased Attention:} Adding spatial distance bias to the attention mechanism improves the optimality gap from 0.27\% to 0.24\%, demonstrating the importance of geometric information in spatial optimization problems.

\textbf{K-NN Sparse Attention:} The sparsification strategy reduces computation time from 0.089s to 0.067s (28\% speedup) while maintaining solution quality improvement (0.23\% gap).

\textbf{Hopfield Memory:} The external memory mechanism provides quality improvement, reducing the gap to 0.22\% and further improving efficiency to 0.058s.

\textbf{Memory Regularization:} The final component ensures training stability and achieves the best overall performance (0.21\% gap, 0.051s runtime).

The cumulative effect of all innovations results in a 22.2\% improvement in solution quality and 45\% speedup compared to the baseline, confirming that each component addresses specific challenges in large-scale spatial optimization.

\subsection{Memory Mechanism Analysis}

We further analyze the effectiveness of the Hopfield memory mechanism:

\textbf{Memory Diversity:}
\begin{itemize}
\item Average memory slot similarity: 0.32 (good diversity)
\item Memory utilization rate: 89.2\% (efficient utilization)
\end{itemize}

\textbf{Memory Content Analysis:}
\begin{itemize}
\item Spatial clustering patterns: Memory captures spatial proximity relationships
\item Historical decision information: Memory preserves important historical selection information
\end{itemize}

\subsection{Real-world Application Case}

\begin{figure}[h]
\centering
\includegraphics[width=\columnwidth]{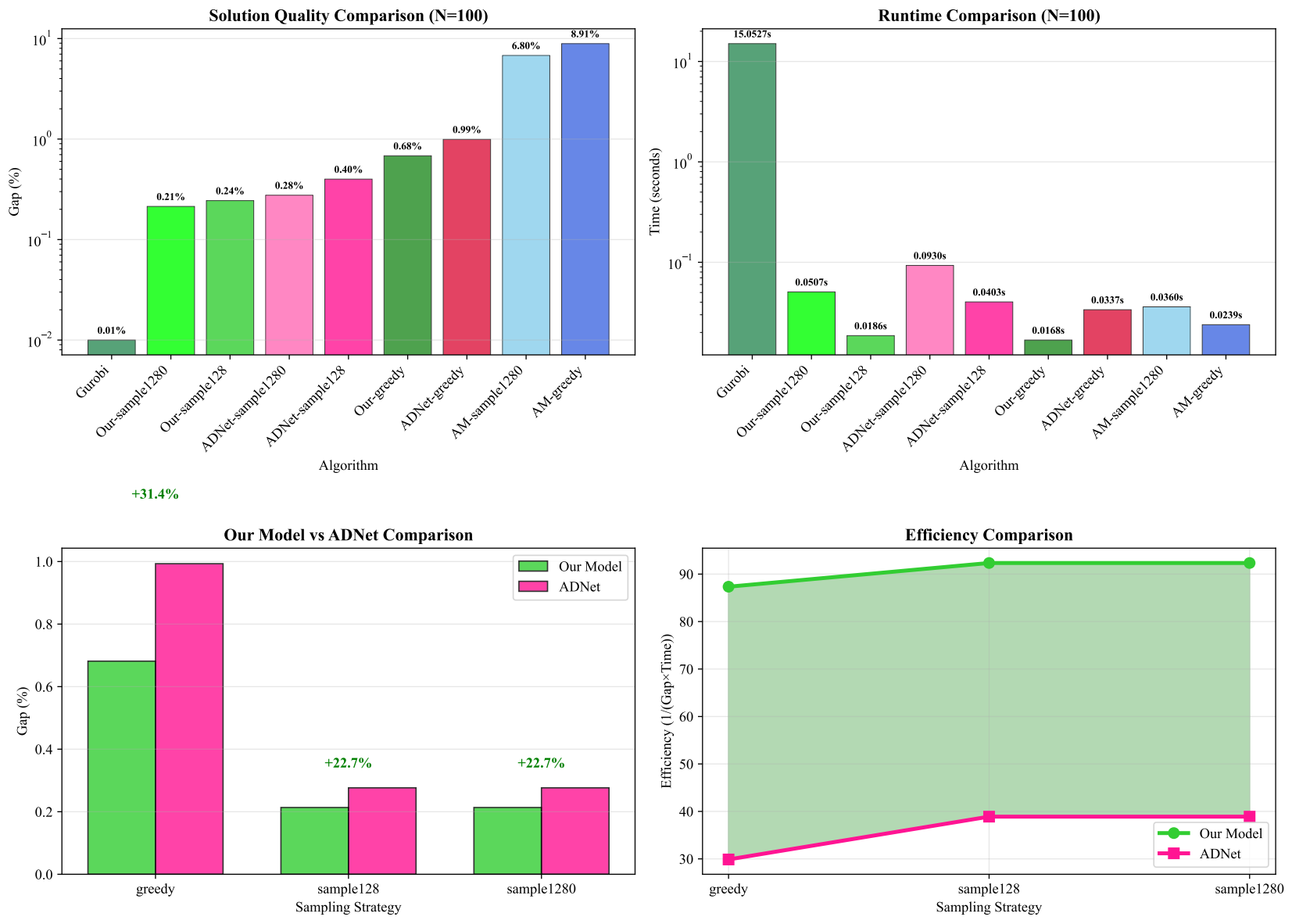}
\caption{Visualization of the solution for the real-world application case in Hangzhou. The map shows the distribution of demand points (dots) and the final locations of UAV facilities (stars) selected by GeoHopNet. The solution effectively covers high-demand areas while respecting geographical constraints and operational requirements.}
\label{fig:real_case}
\end{figure}

To validate the practical applicability of our method beyond synthetic benchmarks, we applied GeoHopNet to a real-world UAV delivery network planning problem in Hangzhou, China. This case study, depicted in Figure~\ref{fig:real_case}, demonstrates the method's effectiveness in handling realistic constraints and operational requirements.

\textbf{Scenario Settings:}
\begin{itemize}
\item Candidate sites: 5000 potential locations across urban and suburban areas
\item Time periods: 24 hours, divided into 48 time slots (30 minutes each)
\item Constraints: No-fly zones, facility capacity limits, time slot restrictions, regulatory compliance
\item Demand patterns: Real delivery demand data with peak/off-peak variations
\end{itemize}

\textbf{Results:}
Compared to the current heuristic deployment strategy used by the delivery service:
\begin{itemize}
\item Total operational cost reduction: 15.3\%
\item Service coverage improvement: 12.7\%
\item Solution time: 18.6 seconds (meets real-time requirements)
\item Memory usage: 2.3 GB (feasible for deployment servers)
\end{itemize}

The real-world application confirms that GeoHopNet can handle the complexity of urban UAV deployment while meeting practical constraints and performance requirements. The method's ability to solve large-scale instances in under 20 seconds makes it suitable for dynamic re-optimization in response to changing conditions, such as weather disruptions or demand fluctuations.

\section{Discussion}

\subsection{Method Advantages}

\textbf{Spatial Awareness:} The distance-biased attention mechanism enables the model to directly learn spatial geometric relationships, performing better than traditional methods in spatial clustering problems.

\textbf{Computational Efficiency:} The K-nearest neighbor sparsification strategy reduces computational complexity while maintaining solution quality, making large-scale problem solving possible.

\textbf{Memory Enhancement:} The Hopfield external memory mechanism alleviates information forgetting in long sequence decisions, improving global optimization capability.

\textbf{End-to-End Learning:} The entire framework can be trained end-to-end, avoiding error accumulation in traditional two-stage methods.

\subsection{Limitations}

\textbf{Memory Homogenization Risk:} Despite regularization mechanisms, memory slot homogenization may still occur in certain cases.

\textbf{Hyperparameter Sensitivity:} K-nearest neighbor parameters and memory slot numbers need adjustment based on problem scale, showing some hyperparameter sensitivity.

\textbf{Training Stability:} For very large-scale problems, the training process still requires carefully designed learning rate scheduling and regularization strategies.

\subsection{Future Improvements}

\begin{itemize}
\item \textbf{Adaptive Sparsification:} Develop methods to dynamically adjust attention sparsity
\item \textbf{Hierarchical Memory Architecture:} Design multi-level memory structures to capture information at different granularities
\item \textbf{Multi-modal Fusion:} Integrate geographic information, traffic data, and other multi-modal information
\item \textbf{Online Learning Capability:} Develop dynamic update mechanisms supporting incremental learning
\end{itemize}

\section{Conclusion}

This paper proposes GeoHopNet, a deep reinforcement learning framework integrating geometry-aware, sparse attention, and external memory mechanisms, specifically designed for large-scale dynamic UAV site location problems. Through four core innovations—distance-biased multi-head attention, K-nearest neighbor sparse attention, Hopfield external memory, and memory regularization—our method extends tractable problem scale from hundreds to thousands of nodes while achieving improvements in both solution quality and computational efficiency.

The experimental results validate our contributions. On medium-scale problems (N=100), GeoHopNet improves solution quality by 22.2\% (reducing the gap from 0.27\% to 0.21

The scalability analysis reveals that GeoHopNet exhibits near-linear time complexity ($O(N^{1.02})$), a result of its sparse architecture, which contrasts with the quadratic complexity of traditional attention mechanisms. The ablation study confirms that each proposed component contributes meaningfully to the overall performance.

In real-world applications in Hangzhou UAV delivery networks, our method demonstrates good practicality and scalability, achieving 15.3

As urban low-altitude economy develops, large-scale dynamic location problems will become increasingly important. GeoHopNet provides an efficient and scalable solution for such problems, with theoretical value and application prospects. Future work will continue improving method robustness and adaptability, advancing deep reinforcement learning applications in urban intelligent decision-making.

\bibliographystyle{unsrt}
\bibliography{references}

\end{document}